# Flashlight Search Medial Axis: A Pixel-Free Pore-Network Extraction Algorithm

Jie Liu[1], Tao Zhang[1*], Shuyu Sun[1*]

1 Physics Science and Engineering, King Abdullah University of Science and Technology, Thuwal, Saudi Arabia, 23955-6900

[*]Corresponding Author: tao.zhang.1@kaust.edu.sa, shuyu.sun@kaust.edu.sa

**ABSTRACT**

Pore-network models (PNMs) have become an important tool in the study of fluid flow in porous media over the last few decades, and the accuracy of their results highly depends on the extraction of pore networks. Traditional methods of pore-network extraction are based on pixels and require images with high quality. Here, a pixel-free method called the flashlight search medial axis (FSMA) algorithm is proposed for pore-network extraction in a continuous space. The search domain in a two-dimensional space is a line, whereas a surface domain is searched in a three-dimensional scenario. Thus, the FSMA algorithm follows the dimensionality reduction idea; the medial axis can be identified using only a few points instead of calculating every point in the void space. In this way, computational complexity of this method is greatly reduced compared to that of traditional pixel-based extraction methods, thus enabling large-scale pore-network extraction. Based on cases featuring two- and three-dimensional porous media, the FSMA algorithm performs well regardless of the topological structure of the pore network or the positions of the pore and throat centers. This algorithm can also be used to examine both closed- and open-boundary cases. Finally, the FSMA algorithm can search dead-end pores, which is of great significance in the study of multiphase flow in porous media.

## 1 Introduction

Porous media have been considered effective fluid carriers in various engineering fields in the last few decades, such as hydrogen and carbon dioxide storage [1, 2], battery design [3], and digital rock study in reservoirs [4, 5]. Porous media have complex topological structures induced by various pores and throats [6, 7], thus, it is imperative that porous media should be accurately described to conduct further numerical investigation.

Porous media are commonly found in a wide range of areas, especially petroleum and natural gas reservoirs [8]. Advancements in CT scanning have enabled the visualization of the inner void structures of rock matrixes [9, 10], which reveals complex topological structures comprising pores and throats. Problems in porous media have been studied using several numerical approaches, such as the lattice Boltzmann method [11], phase field method [12], and smoothed-particle hydrodynamics [13-15]. Molecular simulation methods have also been employed to reveal the microscale mechanisms [16-22]. However, calculating single- or multiphase flow in complex pore spaces is usually computationally challenging. Numerical discretization methods, such as finite volume and finite element methods, require the generation of highly accurate unstructured meshes for porous media [23, 24]. Therefore, a simplified numerical method is necessary to solve problems in complex porous media.

The first pore-network model (PNM) was proposed by Fatt in 1956 [25]; a regular two-dimensional lattice was built using random radii to predict the capillary pressure and relative permeability. However, a regular pore network is not capable of representing the topology and geometry of porous media. Bryant et al. studied pore-network extraction from a realistic porous media in 1992 [26], in which uniform spheres were packed, leading to the conclusion that the pores were packed in a tetrahedral configuration. Furthermore, they predicted the relative permeability of a sand pack, which agreed well with results for sandstone. The growth of PNMs has enabled them to overcome the problems associated with irregular lattices, varying wetting conditions, and multiphase flow [27-29]. This technique is also useful for handling more complex problems, such as non-Newtonian, non-Darcy, and reactive flows [30, 31]. The computational speed of PNMs is considerably faster than that of mesh-based methods; however, their accuracy relies on pore-network extraction.

The medial axis and maximal ball methods are commonly used to extract the pore networks of porous media [32, 33]. Based on the inherent porous characteristics of a porous medium, the medial axis is situated in the middle of the geometric pore space [34]. A significant advantage of the medial axis method is its ability to reduce dimensionality. For instance, in the case of the two-dimensional domain, the medial axis is comprised of medial lines, while in the case of the three-dimensional domain, the medial axis is comprised of the intersection of several medial surfaces. Burning algorithms are commonly used to determine the medial pixel or voxel in the medial axis method. This process involves burning the cells from the solid phase to the void phase one layer at a time [35]. In the maximal ball method [36], the largest inscribed balls are searched in the void space, and balls contained within other balls are removed. The maximal balls are used to identify pores, and the smallest balls are used to identify throats. However, in both the medial axis and maximal ball algorithms, extraction relies on image quality, which is typically based on the resolution. Thus, a single extraction algorithm may yield varying results when subjected to varying resolutions.

In this study, we propose the flashlight search medial axis (FSMA) algorithm, a pixel-free pore-network extraction algorithm for continuous spaces. Theoretically, the FSMA algorithm has a high pore-network extraction efficiency, as dimensionality-reduced search is performed in a two-dimensional or three-dimensional space, where only a small part of the data are used to identify the medial axis.

## 2 Methodology

### 2.1 Characteristics of medial axis

The minimum distance between a random point in the void space and a point in the solid phase is determined as follows:

$$dist(x, D) = \min\{dist(x, y), y \in D\}, \quad (1)$$

where *D* represents the solid phase, *y* represents the point in the solid phase, *x* represents the point in the void phase, and *dist* represents the minimum distance. In an image, a pixel of a solid phase can be visualized and considered a point in that solid phase. The distance map is obtained using the above equation. Fig. 1 shows that the *dist* of the pore center is the maximum in that pore space, which is similar to the mountain top. The points on the medial axis are local maxima; the gradient of *dist* is discontinued, and the curve resembles a mountain ridge. One special point is the throat center, where *dist* is the minimum along the medial axis. The throat center is also a saddle point; the minimum and local maximum are in the direction of the medial axis and the direction perpendicular to the medial axis, respectively. The medial axis comprises the discontinuation points of *dist*'s gradient, which are called critical points, suggesting that the medial axis can be determined accordingly. As depicted in Fig. 1, the total length between two pores is defined as the summation of two parts of lengths from the pore center to the throat center, which is expressed as follows:

$$L_t = L_1 + L_2, \quad (2)$$

where $L_t$ is the length between the two pore centers and $L_1$ and $L_2$ represent the lengths between the pore centers and the throat center. If the inscribed sphere is considered the pore body, then the following equation is the simplest way to determine the length of the throat channel:

$$L_{throat} = L_1 - R_1 + L_2 - R_2, \quad (3)$$

where $L_{throat}$ is the length of the throat channel and $R_1$ and $R_2$ are the radii of two inscribed spheres. However, the volume of the pore body is larger than the volume occupied by the inscribed sphere. The length fraction between the pore body and the throat channel is handled as follows:

$$L_{throat} = L_1 - \alpha \frac{L_1 W_{throat}}{R_1} + L_2 - \alpha \frac{L_2 W_{throat}}{R_2}, \quad (4)$$

where $W_{throat}$ is the *dist* value at the throat center and $\alpha$ is the pore–throat segmentation coefficient. In this regard, the segment coefficient is usually set to 0.5 or 0.6 in previous studies [36, 37].

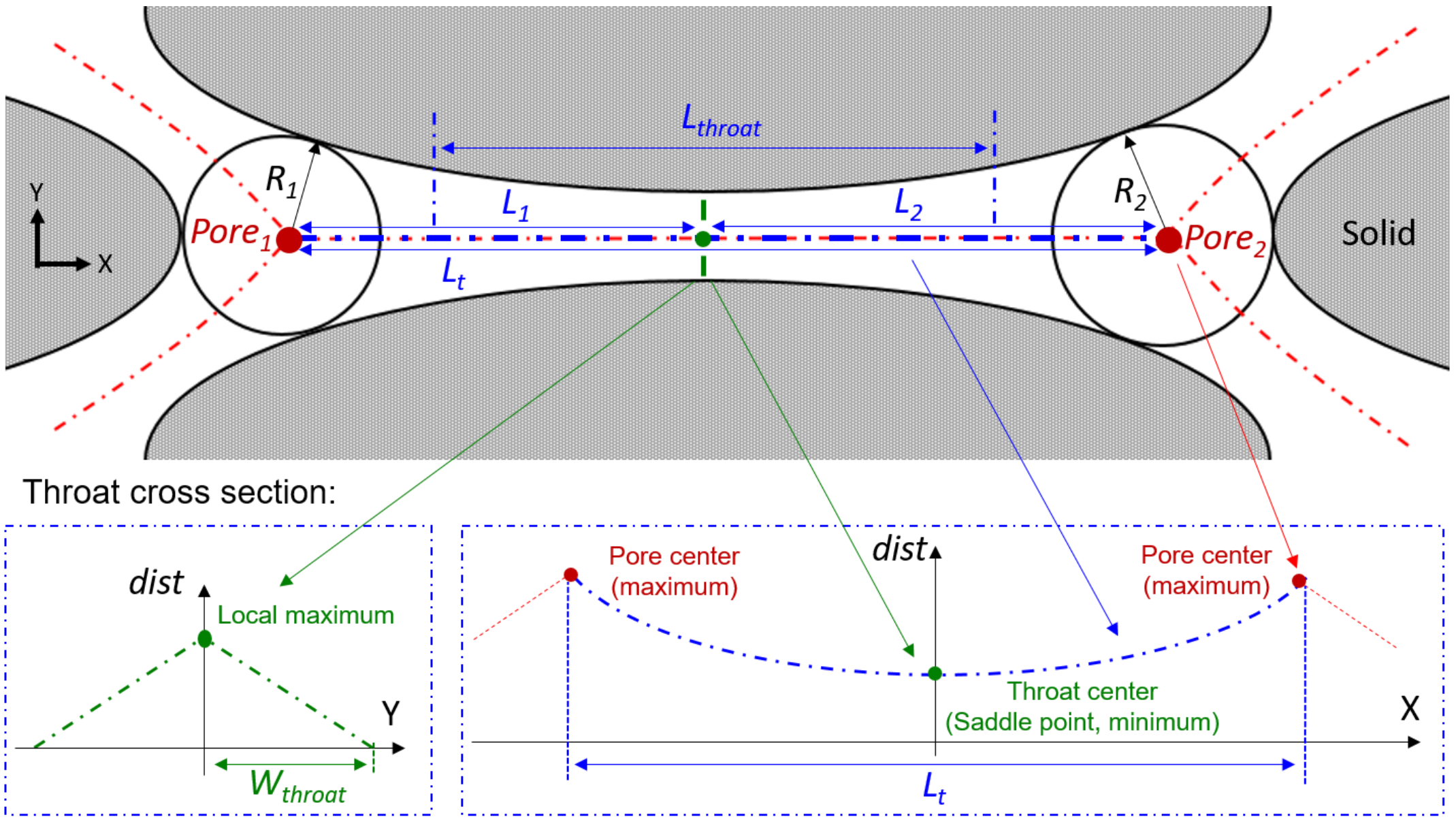

Fig. 1 Schematic of medial axis characteristics.

### 2.2 FSMA pore-network extraction algorithm

#### 2.2.1 Two-dimensional pore-network extraction algorithm

Based on the properties presented in Section 2.1, two-dimensional pore-network extraction is completed as follows:

Step 1: Determine the pore center from a random point in the void space using the steepest-descent method. The calculation domain for the initial point is discretized in the form of a circle to search for the maximum descent. The search range should be as short as possible to ensure the accuracy of pore-center calculation. In our experience, this range should be equivalent to the size of a pixel. Thus, the point is updated with the search step size of the range until the pore center is reached. This process can be likened to a person climbing the steepest route to reach a mountain's summit.

Step 2: From the first pore center, find the medial axis directions. On the position of pore center, the *dist*(*x, D*) is maximum, so there are some critical points lie around the pore center. Circle discretization is applied around the pore center. The medial axis can be determined via a critical point. As seen in Fig. 2, in a case involving three medial axes from one pore center, three critical points are determined using the gradient and Laplace of *dist*.

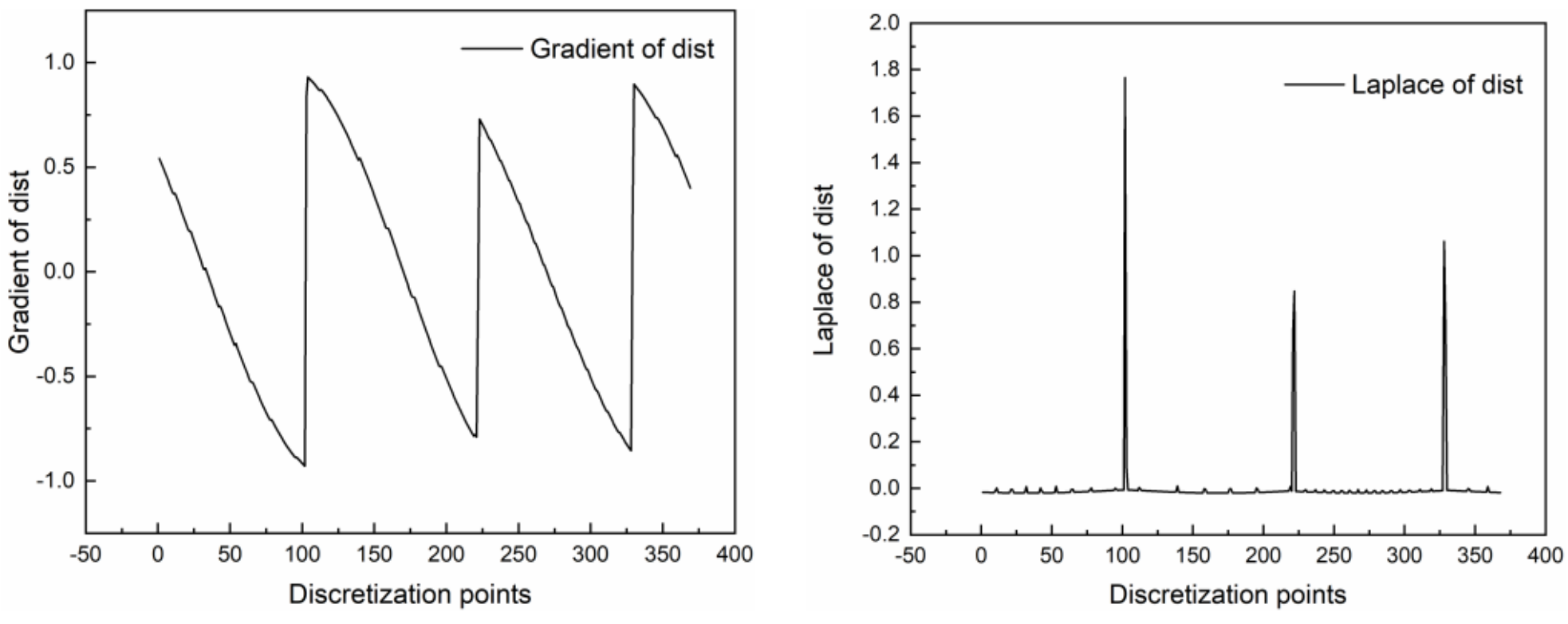

(a) (b)

Fig. 2 (a) Gradient of *dist* and (b) Laplace of *dist*.

Step 3: Flashlight searching method to update the critical point on the medial axis. Once the direction of the medial axis is determined, a fan-shaped region is adopted to search for the next critical point. The radio of the fan-shaped region should be smaller than the radio of the circular search region. The critical point can be updated along the medial axis by following this rule. As shown in Fig. 3, the medial axis (*ma*) is determined one step at a time, and the critical point (*cp*) at each step is searched by building a fan-shaped region with a radius of $r_s$. Thus, the total length between two pores ($pc_1$ and $pc_2$) is expressed as follows:

$$L_t = L_1 + L_2 = \sum_1^{n_1} r_{si} + \sum_{n_1+1}^{n_2} r_{si}, \quad (5)$$

where $r_{si}$ is the search radius at step $i$, $n_1$ is the step number from the initial pore center to the throat center, and $n_2$ is the step number from the throat center to the next pore center.

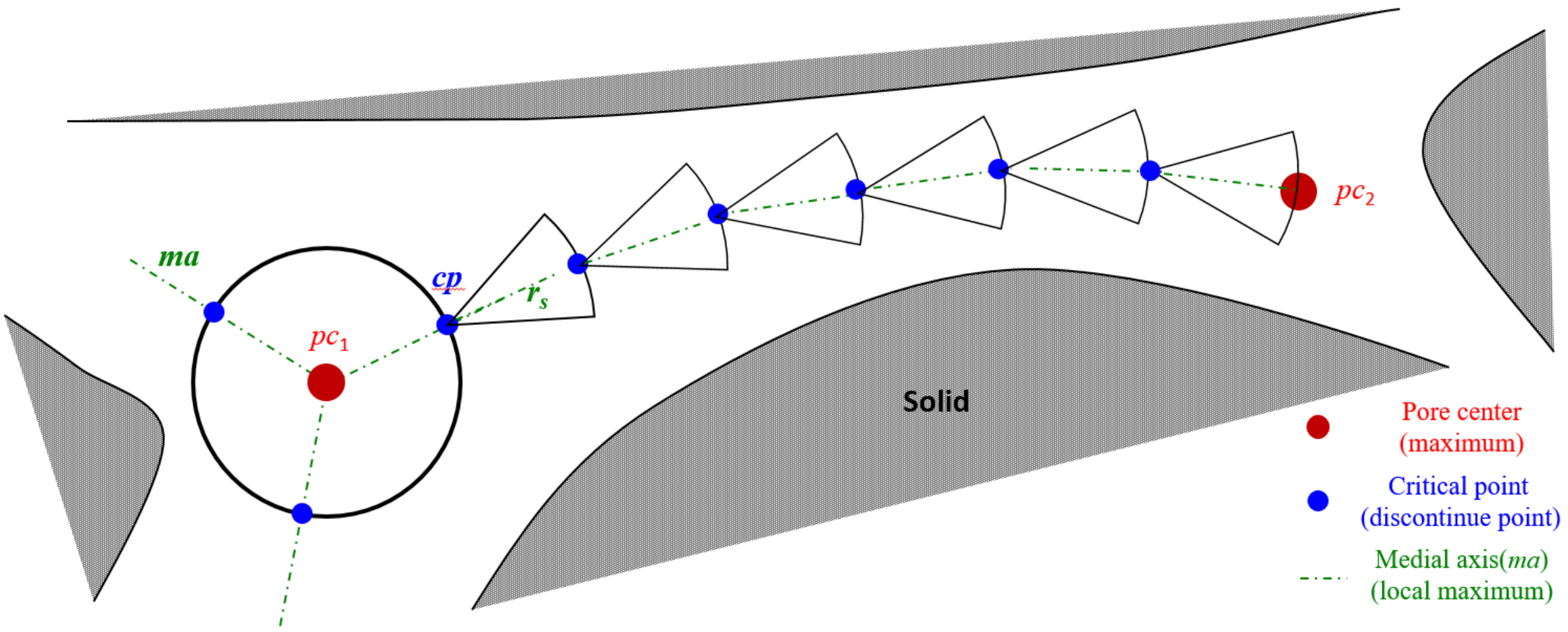

Fig. 3 Schematic of two-dimensional flashlight search.

Step 4: Judge the next pore center and throat center. The critical point is updated via the FSMA method. If the *dist* on the critical point reaches its maximum value, then that critical point is the pore center. If the *dist* on the critical point reaches its minimum value, then that critical point is the throat center.

Step 5: Exclude the overlapped pore centers. Since the medial axis will be determined from one pore center to another pore center, there will be some overlapped pore centers. Thus the overlapped pore centers need to be excluded to avoid repetitive calculations.

### 2.2.2 Three-dimensional pore-network extraction algorithm

The idea underlying three-dimensional pore-network extraction is similar to the above, but the three-dimensional space introduces greater complexity.

Step 1: Find a pore center in the three-dimensional void space by using the steepest-descent method. Unlike the circular search region in the two-dimensional case, a spherical search region is adopted in the three-dimensional study, as seen in Fig. 4(a). The rest of the settings are the same as those in the two-dimensional case.

Step 2: Find the medial axis directions from the first pore center. Three-dimensional discretization is conducted, with the pore center regarded as the sphere center; this is achieved via projection from a cube to a sphere, as shown in Fig. 4(a). The *dist* map on the sphere surface can then be calculated, as seen in Fig. 4(b). Hence, discretization performs better in terms of mesh homogeneity. In Fig. 4, six local maximum points are observed on the sphere surface, which indicates the presence of six medial axis directions. The cross point of the medial surfaces is the medial axis point, suggesting that several medial surfaces form the medial axis.

Compared with classical (spherical-coordinate) discretization, the projection from the cube to the sphere has better discretization performance. Fig. 5 exhibits the *dist* results of both discretization approaches along the medial axis. Fig. 5(b) (projection) shows a smoother curve, which is better for identifying the pore and throat centers. Misidentification may occur in Fig. 5(a), as the rough curve may induce some local minimum values along the medial axis.

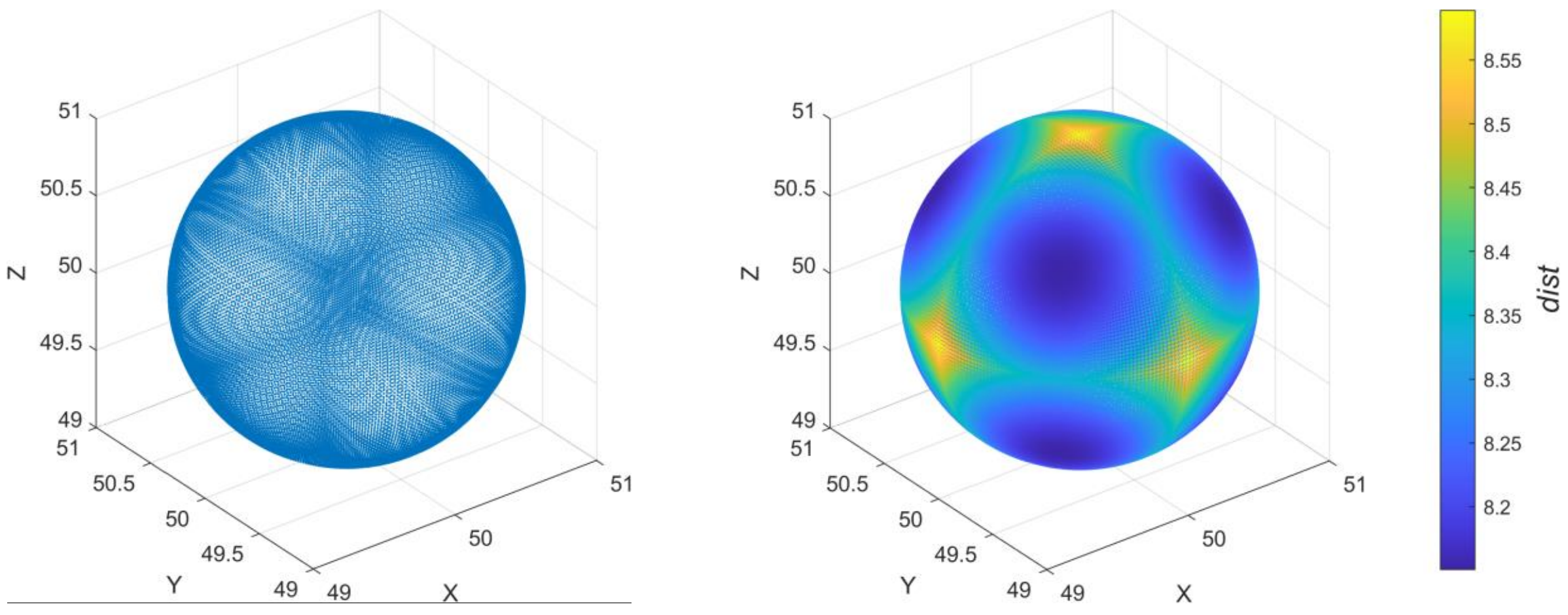

(a) (b)

Fig. 4 Uniform sphere-packing scenario: (a) projection from cube to sphere and (b) *dist* map on sphere surface.

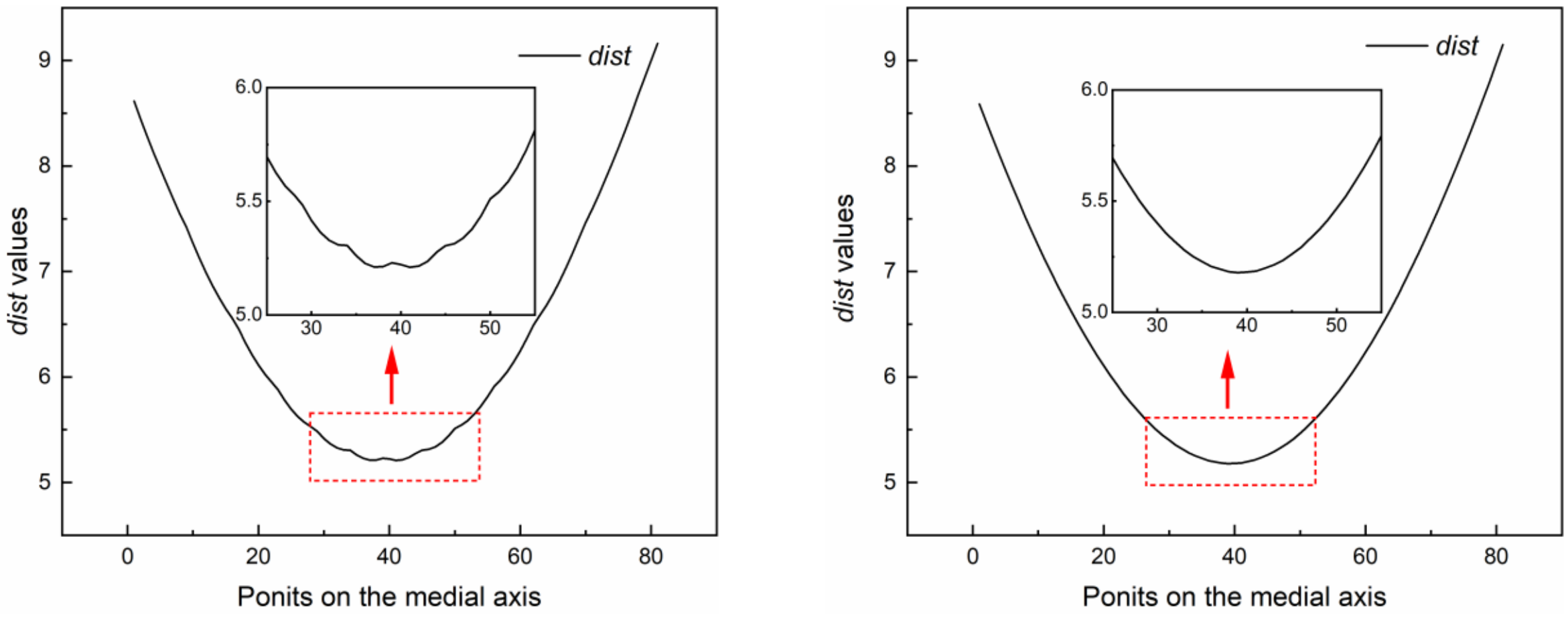

(a) (b)

Fig. 5 *dist* values on medial axis obtained via (a) discretization using spherical coordinates and (b) discretization using cubic projection on sphere surface.

Step 3: Update the critical points on the medial axis via the FSMA algorithm. Unlike the fan-shaped search region in the two-dimensional space, a cone-shaped search region is used in the three-dimensional space. The rest of the settings are the same as those in the two-dimensional case. In this manner, the critical points on the medial axis are determined by identifying the medial surfaces, as shown in Fig. 6.

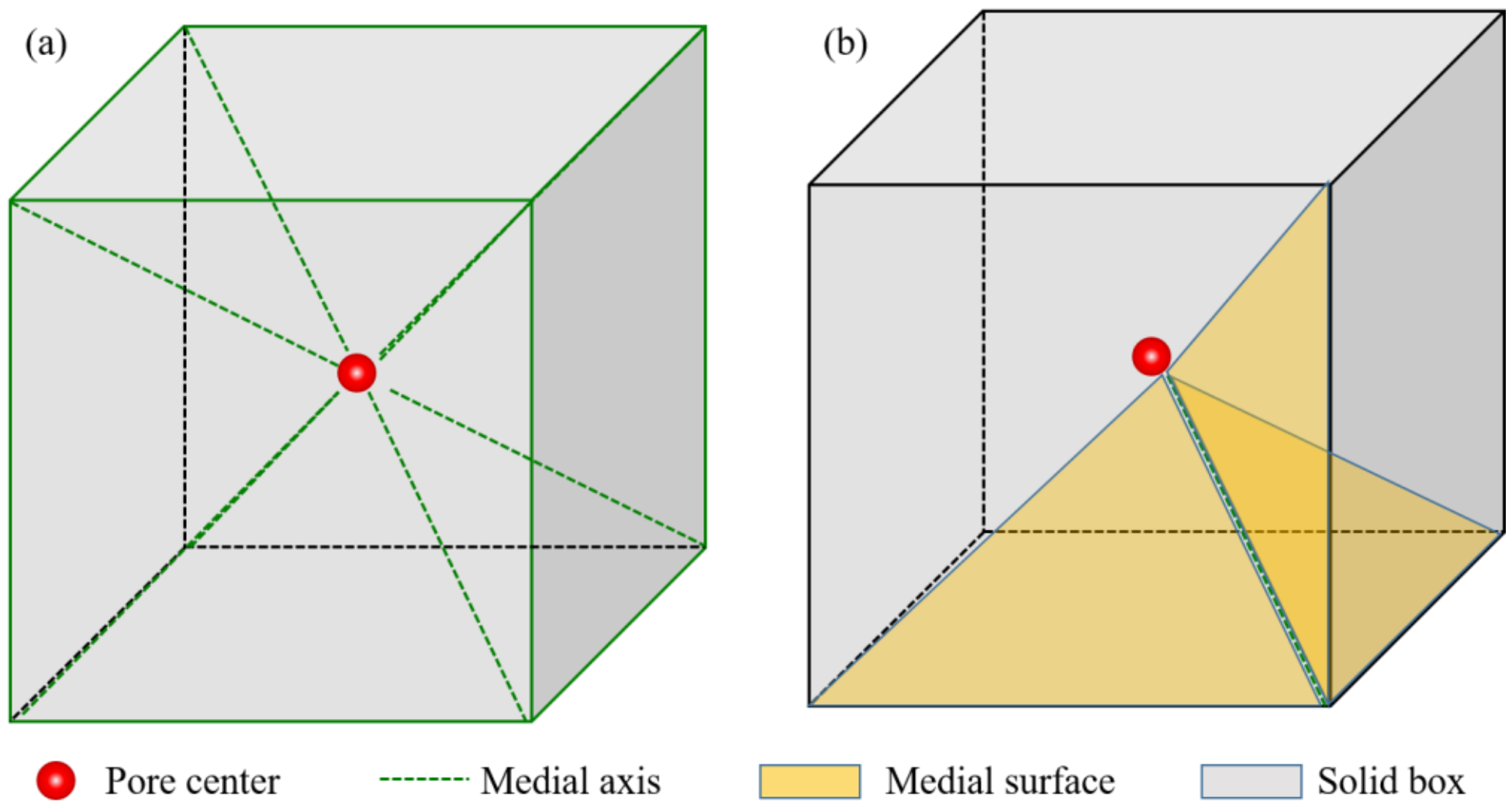

Fig. 6 Schematic of determination of medial axes: (a) eight medial axes from pore center in cubic solid box and (b) one medial axis formed by three medial surfaces.

Step 4: Judge the next pore center and throat centers. In the cone-shaped search region, the medial surface is reduced to a medial axis. Thus, the pore center in the cone-shaped search region is the cross point of several medial axes; the pore center has the maximum *dist* value. Along the medial axis, composed of the critical points, the throat center can be identified on the minimum of *dist*.

Step 5: Exclude the overlapped pore centers. This step is the same as the operation in the two-dimensional scenario.

The pseudocode of the FSMA algorithm is summarized as follows:

**Algorithm** FSMA

```
Input: a random initial point in the void space p; the known solid-phase data S; the
search radius r_s to determine the critical point cp.
Output: the pore center pc; the medial axis ma;
Find the first pore center p_1 using the steepest-descent method; // According to Step 1
for search pore = 1: m
// From new-reached pore centers to further-reached pore centers
   for search step = 1: n // For the calculation of one strip of medial axis
      Discretize the search region;
      Determine the neighboring solid data from the previous cp;
      Calculate the dist value of discretized points;
      Judge the critical point;    // Introduced in Step 2 and 3
      if critical point cp_i reaches the pore center // Judge the pore center, in Step 4
         pc_i = cp_i;
         ma_i = {cp_begin, …, cp_i};
      end
   end
   Update the new-searched pore centers;
   Exclude overlapped pore centers;    // According to Step 5
end
```

### 2.3 Comparative analysis of FSMA and pixel-based algorithms

Regarding to the traditional pore-network extraction methods, such as the maximal ball algorithm, the shape of the inscribed sphere highly depends on the image resolution. If the inscribed sphere is built on a low-resolution image, the boundary information of the inscribed sphere cannot be accurately described, and high-resolution images are computationally expensive. Therefore, the ability of the FSMA, a pixel-free method, to identify the medial axis regardless of image resolution is a natural advantage. Because of the dimensionality reduction idea during the search, computational complexity is reduced compared with that of traditional methods. The computational complexity of the FSMA algorithm is summarized in Table 1. As the steepest-descent method is only used once (for determining the first pore center), its complexity is not considered in the total code. $k$ is the comparison number for the pore and throat centers; this value is usually too small to add to the computational complexity.

Table 1 Computational complexity of FSMA algorithm

| Line | Complexity | Description |
|---|---|---|
| Line 3 | $O(s^2)$ | Steepest-descent algorithm |
| Lines 6–9 | $O(n \log n)$ | Search for critical points |
| Lines 10–13 | $O(k * n)$ | Determination of pore and throat centers |
| Lines 15–16 | $O(n)$ | Updating of pore and throat centers |
| Lines 4–17 | $O(m * (n \log n + k * n + n))$ | |
| Total | $O\left(s^2 + m * (n \log n + k * n + n)\right) = O(mn \log n)$ | |

The computational complexity of traditional pore-network extraction methods is expressed as follows:

$$O = (M + N)\left(\frac{L}{\varepsilon}\right)^d \tag{6}$$

where $O$ represents the computational complexity; $M$ and $N$ are the numbers of pore and throat centers; $d$ represents the dimensionality; $L$ represents the size of the computational domain; and $\varepsilon$ is the discretized unit, which is the size of the pixel or voxel in an image. Hence, $\left(\frac{L}{\varepsilon}\right)$ is the number of points that should be calculated. The computational complexity of the FSMA algorithm is expressed as follows:

$$O = (M + N)\log\left(\frac{L}{\varepsilon}\right). \tag{7}$$

In a case where $\left(\frac{L}{\varepsilon}\right) = 10$ in a three-dimensional space, the complexity of the FSMA algorithm is 1000 lower than that of traditional methods, in which all points need to be considered for one-time calculation.

## 3 Validation and discussions

### 3.1 Pore-network extraction from two-dimensional porous media

Although two-dimensional porous media have few application scenarios, such as artificial microfluidic chips, they should still be used for validation. A two-dimensional porous medium is built in a domain sized 100 × 100. The solid-phase regions are closed using known boundary points, suggesting that the *dist* value in the void phase can be calculated from these solid boundary points. The inner points in the solid phase are unnecessary, which helps save computer memory.

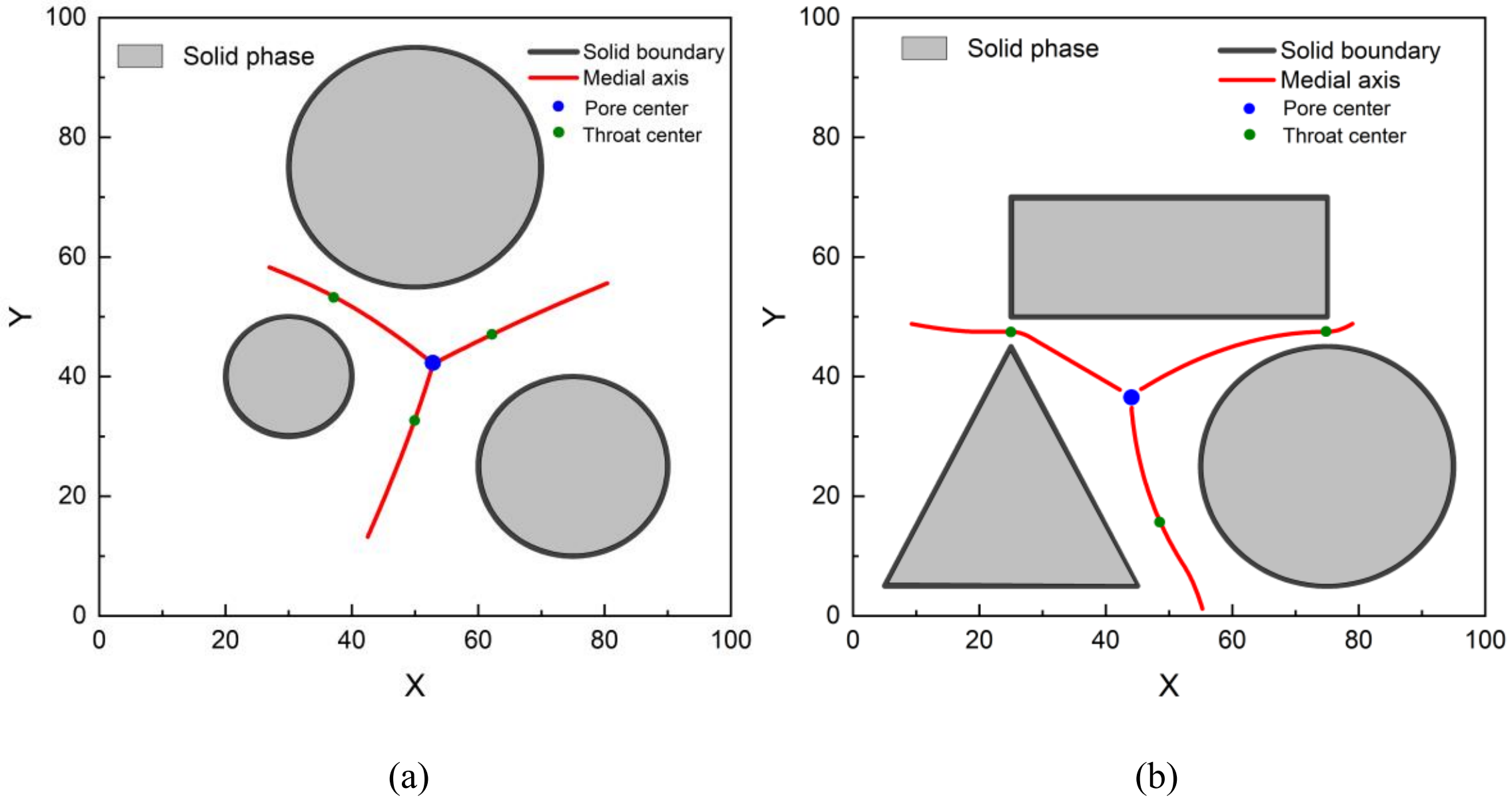

Fig. 7 Pore-network extraction in two-dimensional scenarios constructed using (a) spheres and (b) different shapes.

Fig. 7 depicts the identification of the medial axis from the first pore center, which verifies the validity of the FSMA algorithm for the two-dimensional porous medium. On the medial axes, the throat centers can be determined accordingly; in addition, the throat centers are located at the positions of the minimum distance between two spheres, as shown in Fig. 7(a), which enables a clear assessment of the throat center calculation. In Fig. 7(b), the medial axis and throat center can also be identified, suggesting that the FSMA algorithm can search the medial axis from one pore center via the dimensionality-reduced search method.

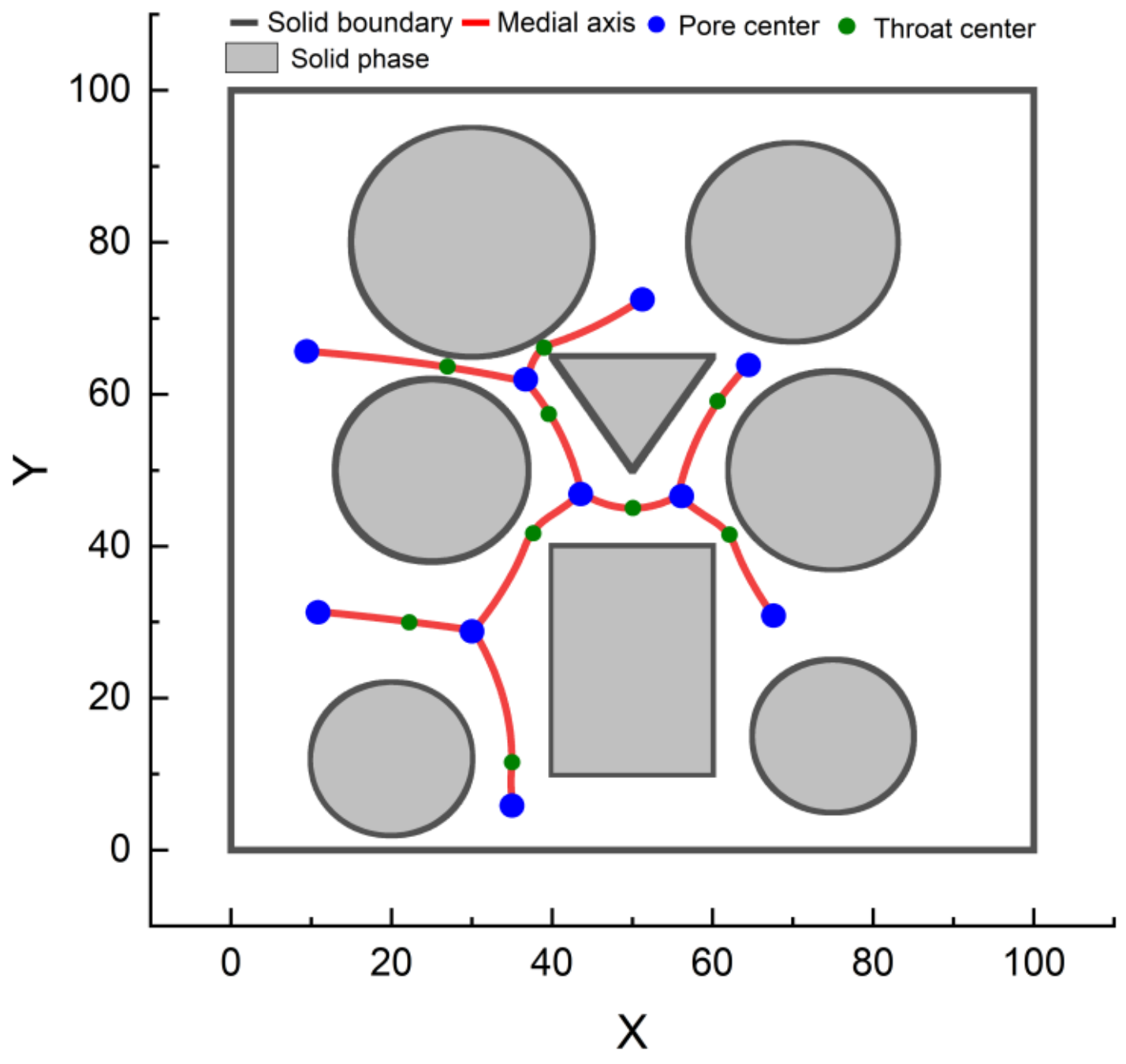

Fig. 8 Pore-network extraction from one pore center to other pore centers in two-dimensional porous medium.

Once the medial axis is determined, the other pore centers can be searched from the first pore center and its medial axes. As shown in Fig. 8, the pore centers and medial axes comprise the topological structure of the porous medium. The pore network is identified gradually, pore center by pore center. Moreover, neighbor search is adopted in the search process. The *dist* values of the discretized points in the fan-shaped search region are calculated within the neighbor range, which is identified from the critical point in the previous step. Therefore, the pore-network construction of the FSMA algorithm only needs a small part of the information in the porous medium.

### 3.2 Boundary treatment

Boundary treatment is critical for pore-network extraction, as PNMs are used for fluid-flow simulation, which is a major application field in petroleum engineering [38, 39].

Inflow and outflow boundaries are necessary for fluid-flow simulation, and dead-end corners are also considered as special boundaries in porous media [40, 41]. The FSMA algorithm has a natural advantage in searching dead-end corners. As the FSMA algorithm is developed by searching medial axes for critical points, a critical point cannot be found in a dead-end corner within a search region. Therefore, dead-end corners are determined, as shown in Fig. 9. Despite the increase in the number of pore centers, the dead-end pore center can be ignored in some scenarios, such as single-phase fluid flow. However, in multiphase fluid-flow problems, dead-end corners play an important role. Fluids with similar wettability to the rock surface are trapped in these corners, thus affecting the subsequent fluid flow.

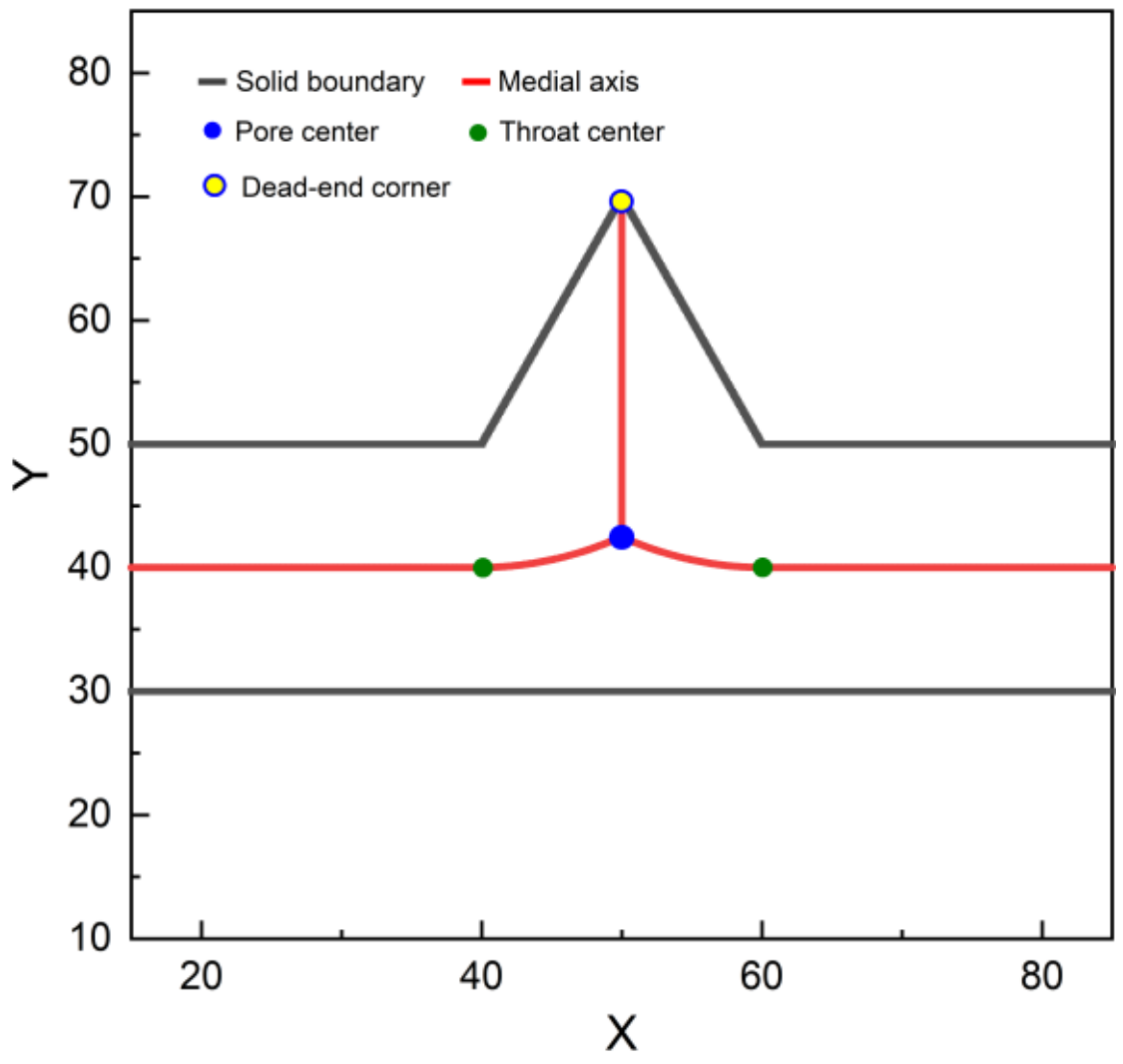

(a)

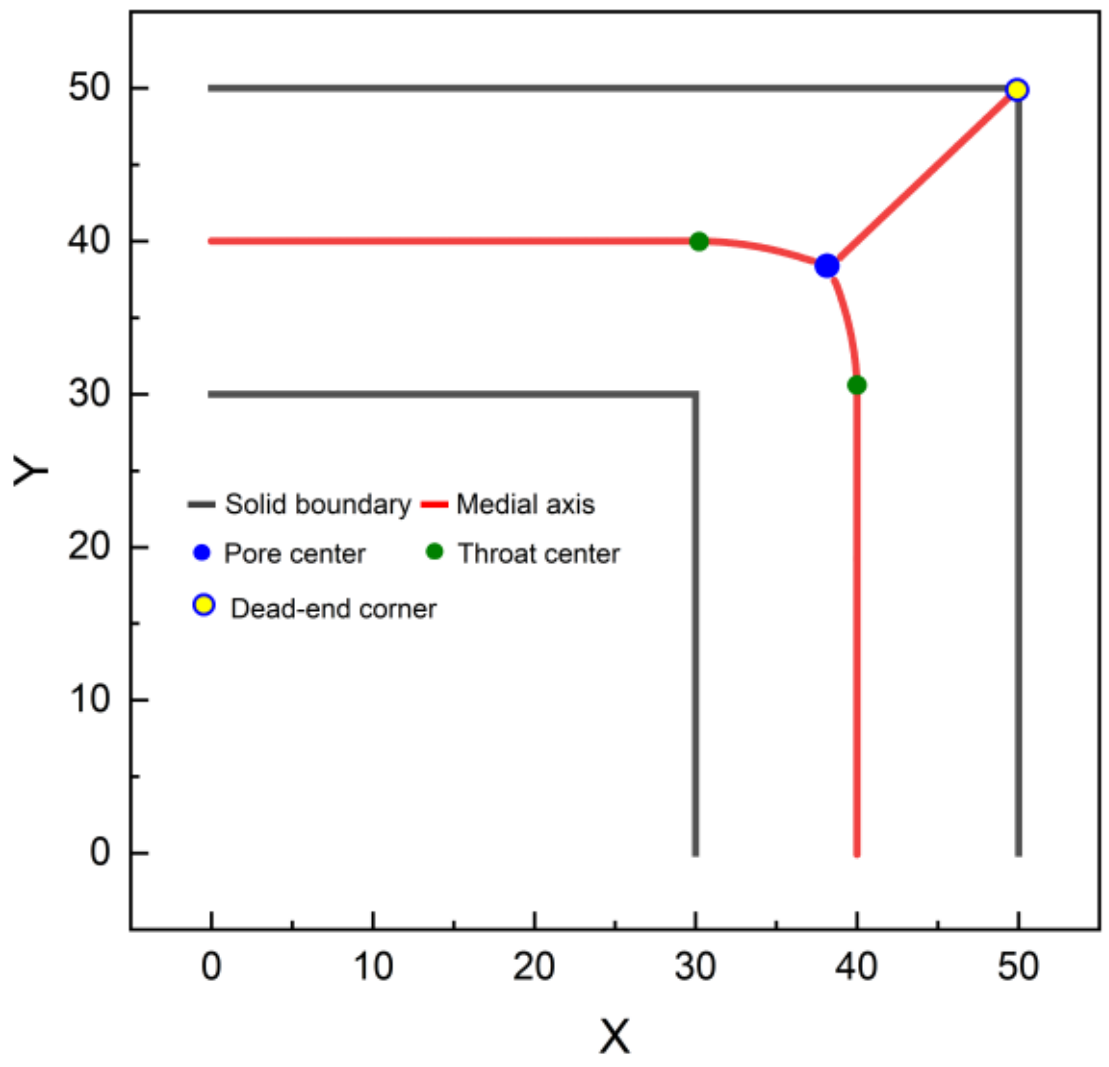

(b)

Fig. 9 Test cases of (a) a straight tube and (b) a elbow tube for FSMA algorithm featuring dead-end corners.

Solid boundaries are a common boundary type in fluid-flow simulation, as the connectivity of pores and throats cannot be confirmed in porous media. In a solid boundary, the surface of a solid phase and a solid box form new pore channels. Dead-end corners are also generated in closed solid boxes, and they can be extracted when necessary. As seen in Fig. 10, the FSMA algorithm can search all critical points and extract the pore network in solid-box domains; this ability is meaningful in the study of reservoir rocks which have bad connectivity.

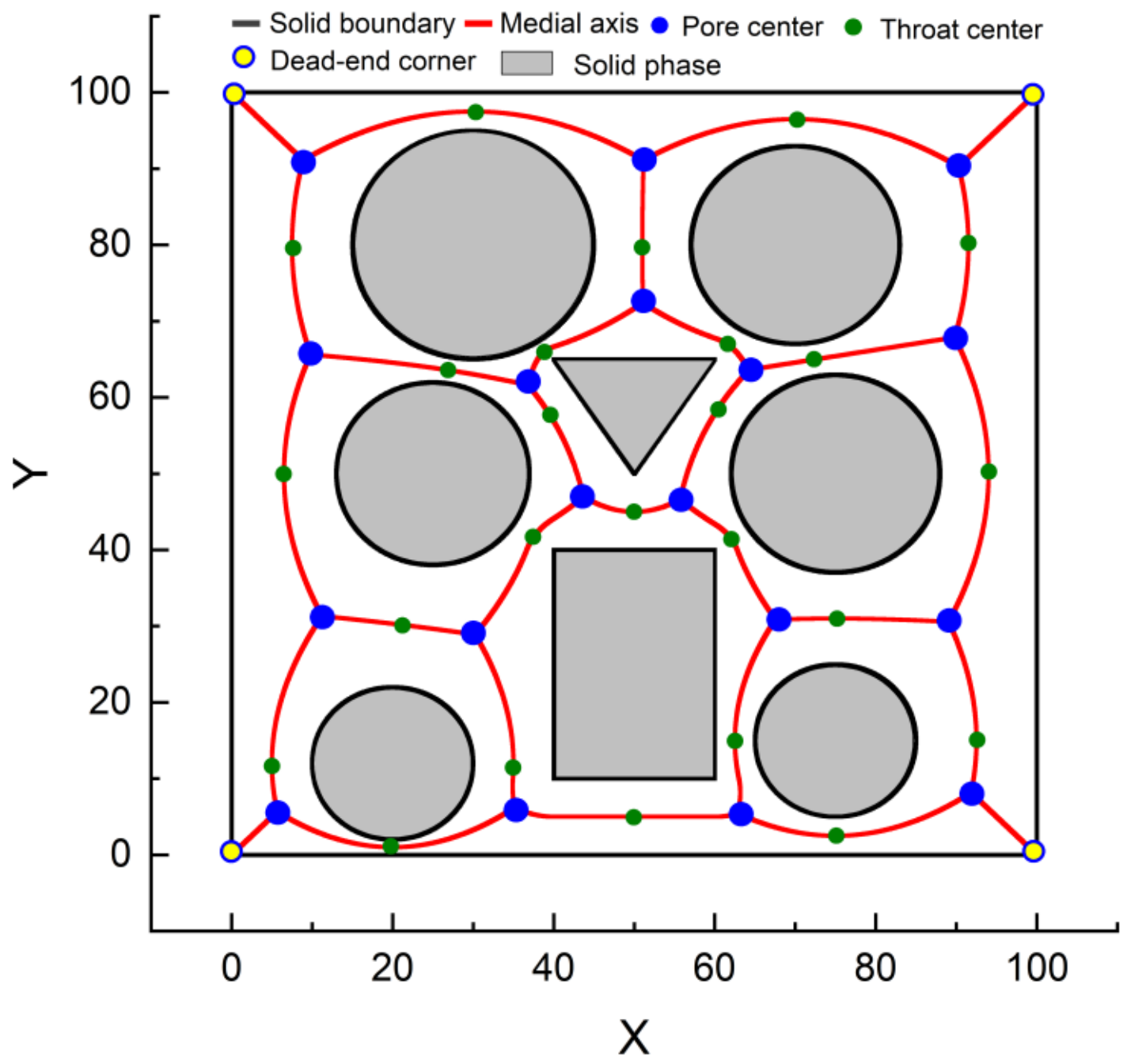

Fig. 10 Pore-network extraction within closed (solid) boundaries via FSMA algorithm.

Open boundaries are also examined using the FSMA algorithm. In cases with open boundaries, the search operation proceeds until the boundary of the computational domain. Fig. 11 illustrates pore-network extraction results in a domain where the right and left sides have open boundaries and the pore centers stop at the boundaries; these are set as the Dirichlet boundary condition and the Neumann boundary condition, respectively [42, 43], and the fluid flow can be realized consequently.

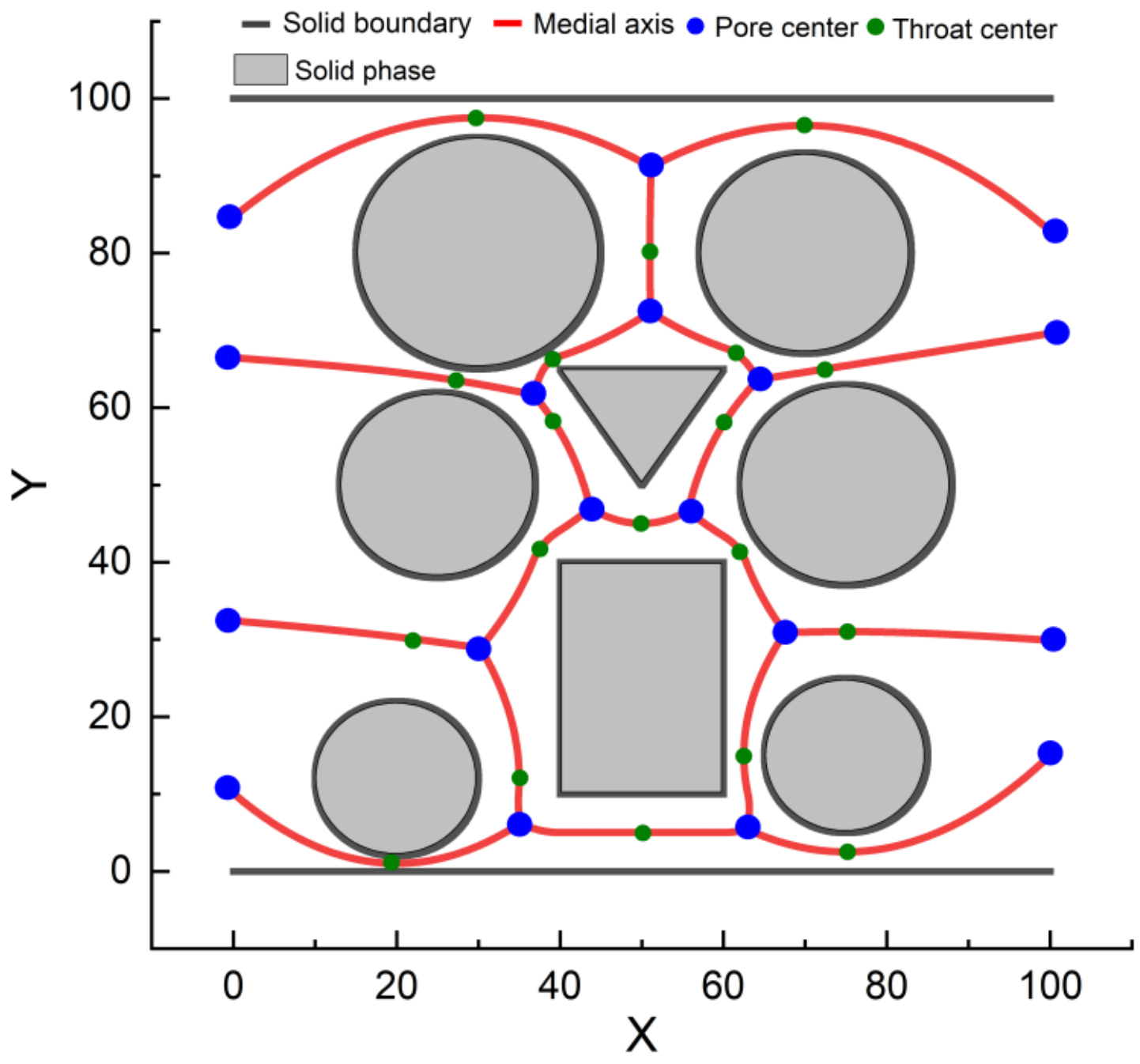

Fig. 11 Pore-network extraction within open boundaries via FSMA algorithm.

### 3.3 Different packing styles for three-dimensional space

As shown in Fig. 12(a), a sphere-packing model is built in a cubic box sized 100 × 100 × 100 to verify the FSMA algorithm in a three-dimensional space. Sixty-four solid spheres are packed in this box in a 4 × 4 × 4 packing style The radius of each sphere is 12.5, suggesting that the spheres are in contact with each other. The surface of a solid sphere is discretized through regular hexahedron projection for better computational performance. The critical points are searched and the medial axis is identified in the three-dimensional space using the computational algorithm described in Section 2.2.2.

A regular packing style is adopted because the medial axis is connected as a regular network and is easily verified. Fig. 12(b) shows the beginning state of the medial axis identity, which is from the first pore center to a neighbor pore center through the connection of the medial axis. As seen in Fig. 12(c), the regular topological structure of the pore network is then extracted, revealing that the FSMA algorithm is applicable to three-dimensional porous media.

Apart from the regular model, as depicted in Fig. 13, an irregular sphere-packing model is used to test the feasibility of the FSMA algorithm in a more complex porous medium, and it performs well.

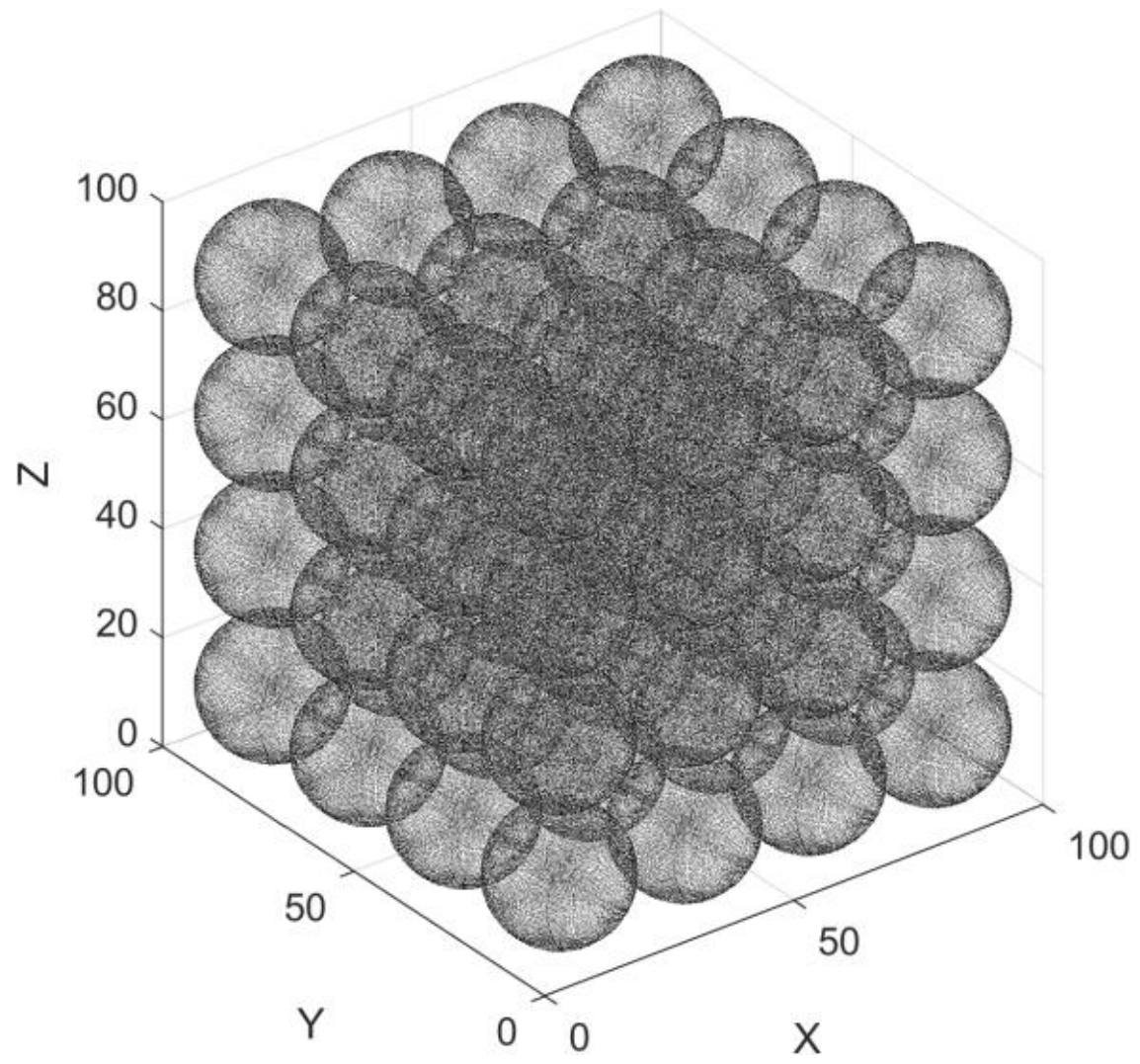

(a)

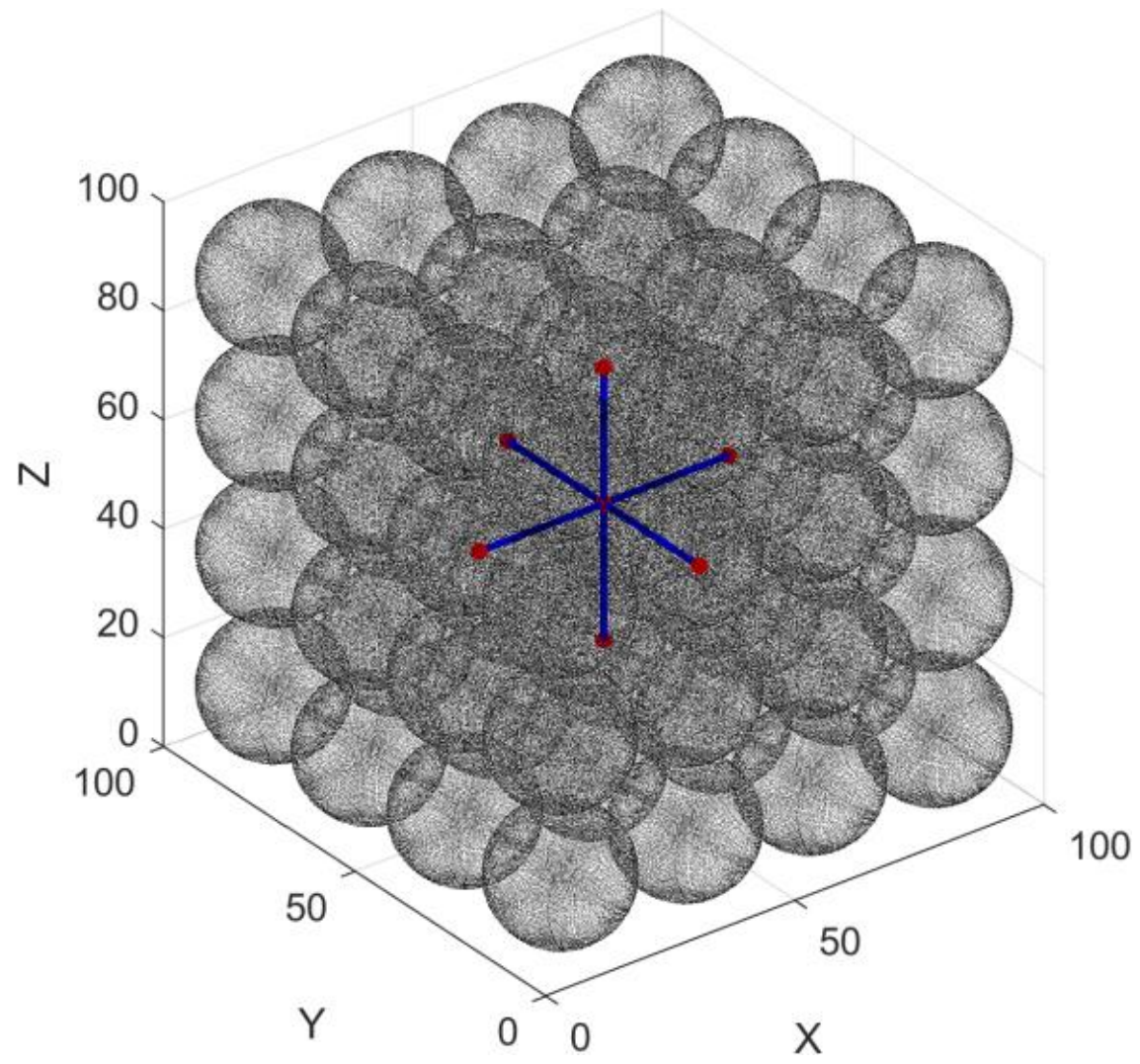

(b)

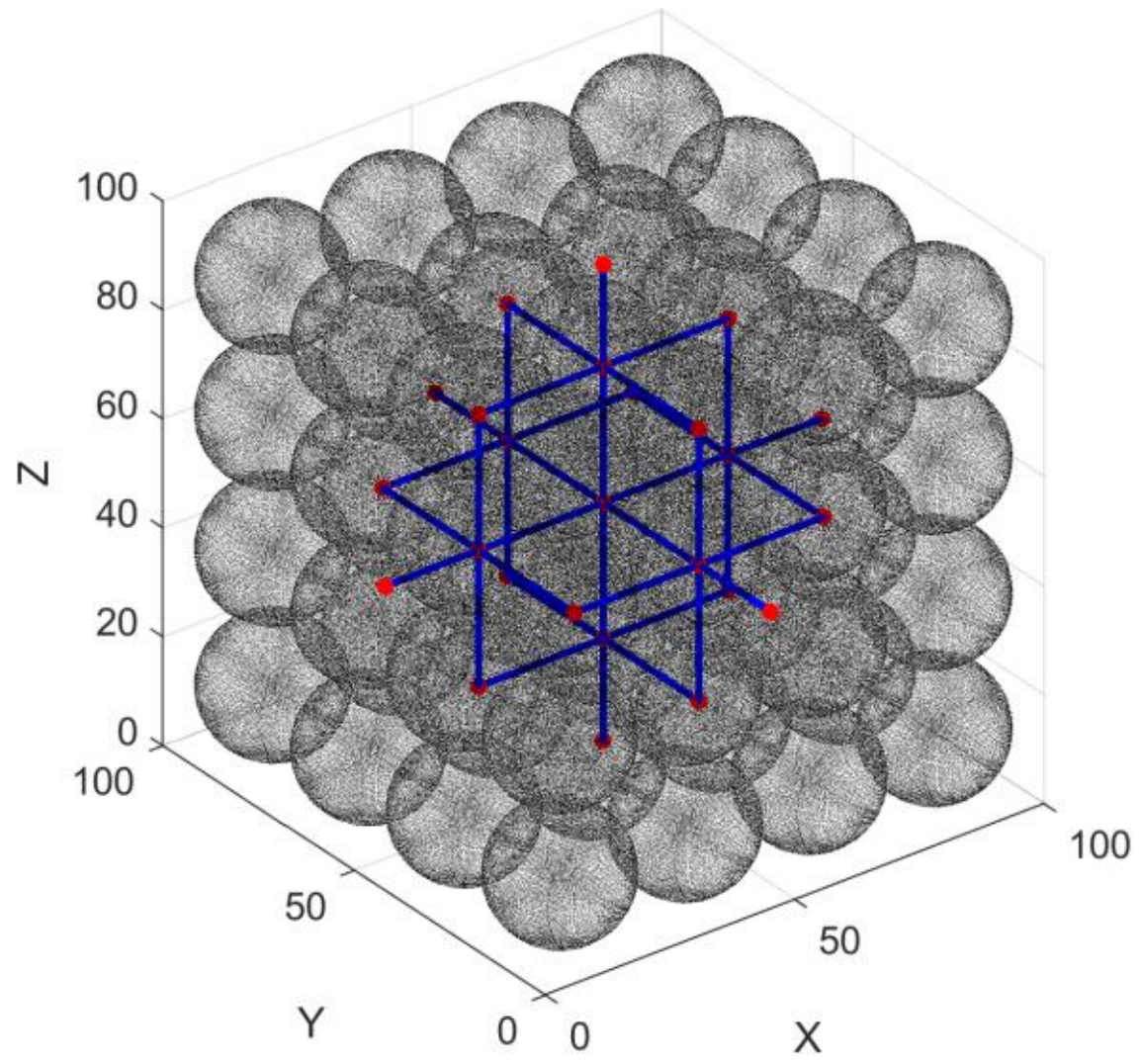

(c)

Fig. 12 (a) Three-dimensional regular sphere-packing model and (b–c) extracted medial axes. Blue line represents the medial axis, and red point represents the pore center.

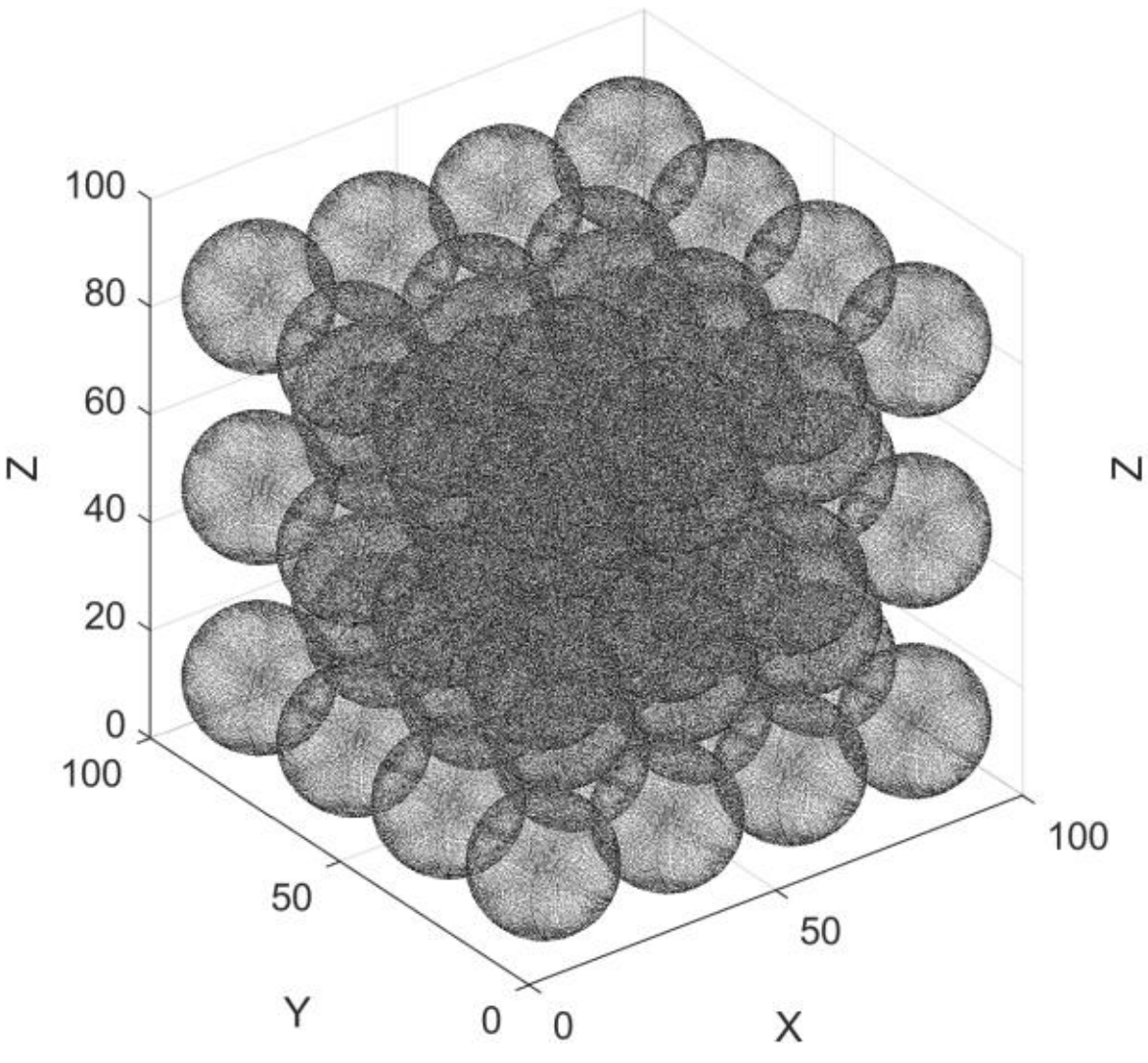

(a)

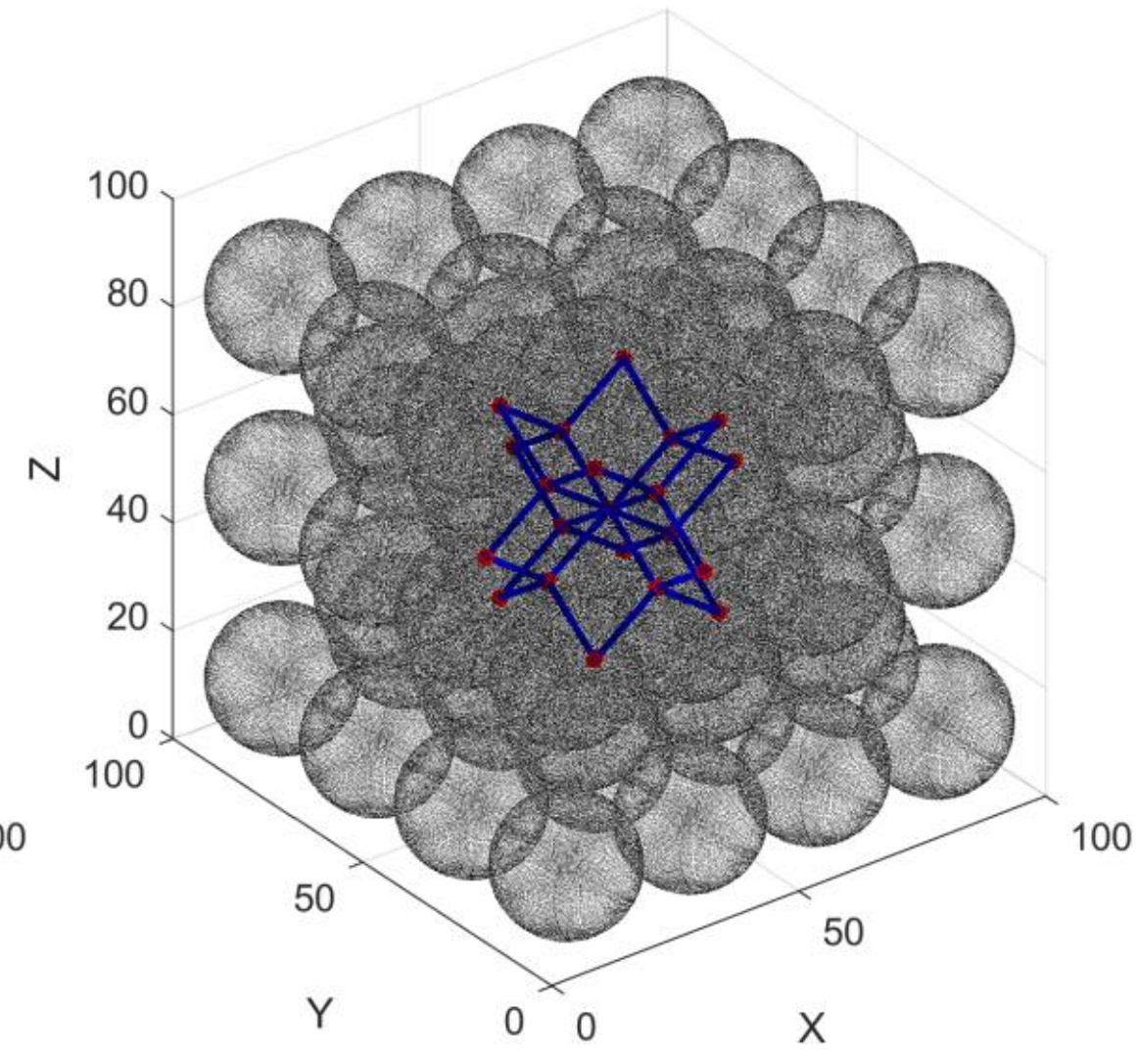

(b)

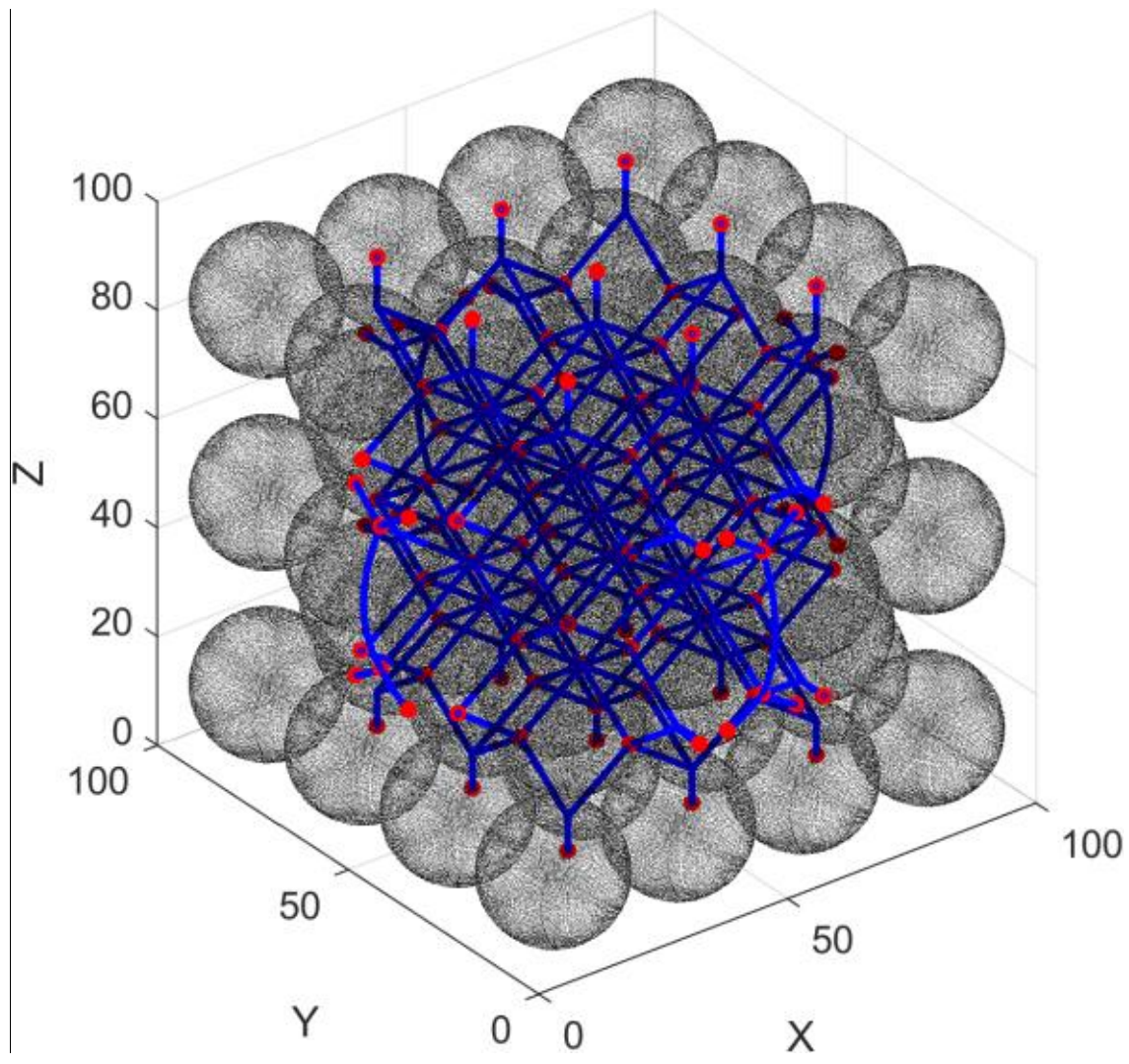

(c)

Fig. 13 (a) Three-dimensional irregular sphere-packing model and (b–c) extracted medial axis. Blue line represents the medial axis, and red point represents the pore center.

## 4 Conclusions

In this study, a pore-network extraction algorithm called FSMA is proposed. The FSMA algorithm is developed in a continuous space rather than a pixel-based space, which makes it a pixel-free method. This algorithm adopts the idea of dimensionality-reduced search, and calculating the global data for one pixel point is unnecessary. Thus, the algorithm has a substantially shorter computational time than traditional methods. In traditional search procedures, each point is calculated in the void space, whereas in the FSMA search procedure, only a few points need to be considered. In addition, the computation is further accelerated using neighbor search. Theoretically, the FSMA algorithm extracts pore networks with low computational complexity and high efficiency. These features are highly advantageous in large-scale pore-network extraction, and this algorithm enables the characterization of the PNM of a reservoir with a representative elementary volume (REV).

Specific computational programs are introduced for two- and three-dimensional spaces, and cases involving two- and three-dimensional porous media are constructed to verify the feasibility of the FSMA algorithm. The results indicate that the topological structure of the pore network is identified and the pore and throat centers are determined accordingly. Furthermore, according to a discussion of the boundary condition for the FSMA algorithm, the algorithm has the natural advantage of being able to search dead-end corners in porous media. Dead-end pore centers are identified by changing the determination conditions of the algorithm for searching critical points, which is useful for studying multiphase flow. Additionally, the algorithm performs well in both closed- and open-boundary tests. In conclusion, the pixel-free pore-network extraction algorithm FSMA can extract pore networks at high theoretical computational speeds and therefore has considerable potential for addressing large-scale problems.

**Acknowledgments**

We would like to express appreciation to the following financial support: National Natural Scientific Foundation of China (Grants No. 51936001), King Abdullah University of Science and Technology (KAUST) through the grants BAS/1/1351-01 and URF/1/5028-01.